%% file: main.tex
\documentclass[
]{ceurart}

\usepackage{listings}
\usepackage{wrapfig}
\usepackage{graphicx}
\usepackage{caption}
\usepackage{floatrow}
\usepackage{subcaption}
\usepackage{hyperref}
\begin{document}

\copyrightyear{2023}
\copyrightclause{Copyright for this paper by its authors.
  Use permitted under Creative Commons License Attribution 4.0
  International (CC BY 4.0).}

\conference{NeSy 2023, 17th International Workshop on Neural-Symbolic Learning and Reasoning, Certosa di Pontignano, Siena, Italy}

\title{A Hybrid System for Systematic Generalization in Simple Arithmetic Problems}

\author[1]{Flavio Petruzzellis}[%
email=flavio.petruzzellis@phd.unipd.it,]
\cormark[1]
\address[1]{Department of Mathematics, University of Padova, Padova, Italy}
\address[2]{Department of General Psychology, University of Padova, Padova, Italy}

\author[1,2]{Alberto Testolin}[%
email=alberto.testolin@unipd.it,
]

\author[1]{Alessandro Sperduti}[%
email=alessandro.sperduti@unipd.it,
]

\cortext[1]{Corresponding author.}

\begin{abstract}
Solving symbolic reasoning problems that require compositionality and systematicity is considered one of the key ingredients of human intelligence. However, symbolic reasoning is still a great challenge for deep learning models, which often cannot generalize the reasoning pattern to out-of-distribution test cases. In this work, we propose a hybrid system capable of solving arithmetic problems that require compositional and systematic reasoning over sequences of symbols. The model acquires such a skill by learning appropriate substitution rules, which are applied iteratively to the input string until the expression is completely resolved. We show that the proposed system can accurately solve nested arithmetical expressions even when trained only on a subset including the simplest cases, significantly outperforming both a sequence-to-sequence model trained end-to-end and a state-of-the-art large language model.
\end{abstract}


\begin{keywords}
  deep learning \sep
  neural networks \sep
  mathematical reasoning \sep
  neuro-symbolic systems \sep
  formula simplification
\end{keywords}

\maketitle

\section{Introduction}
\label{par:intro}
\input{1-intro}

\section{The Proposed Hybrid System}
\label{par:system}
\input{1.5-system}

\section{Methodology}
\label{par:methods}
\input{2-method}

\section{Results}
\label{par:results}
\input{3-results}

\section{Related work}
\label{par:related}
\input{4-related}

\section{Discussion}
\label{par:discussion}
\input{5-discussion}

\bibliography{9-bib}
\clearpage

\section{Appendix}
\label{par:appendix}
\input{6-appendix}
\end{document}

%% file: 1-intro.tex
Designing systems that are able to learn abstract reasoning patterns from data is one of the fundamental challenges in artificial intelligence research.
In particular, the capacity to generalize compositionally \citep{jair_HupkesDMB20}, to reuse learned rules on unseen algorithmic problems that fall outside of the training distribution \citep{mahdavi2023towards}, and more generally to exhibit high-level reasoning abilities are considered long-standing issues by the machine learning community.
One possibility to empower artificial intelligence systems with reasoning abilities is to combine statistical-based approaches, which can flexibly adapt to the properties of the observed data, with logic-based modules, which can be engineered according to a set of predefined rules \cite{hitzler2022neuro}.

In this work, we introduce a novel neuro-symbolic architecture that solves reasoning problems on sequences of symbols in an iterative fashion. Our goal is to build a system that can learn to solve compositional tasks by only observing a subset of problem instances and then generalize the acquired knowledge to out-of-distribution  (possibly more difficult) instances.
The proposed system is made up of two parts: a deep learning module that is trained to generate candidate solutions that simplify a given input sequence, and a deterministic component that combines the original input and the output of the neural module to produce the final output.
Thanks to its hybrid nature, our system can learn the fundamental solution steps for a set of symbolic reasoning problems, and then apply these operations iteratively to novel problem instances.

We evaluate our model on the task of simplifying arithmetic expressions, which is well suited to study systematic reasoning abilities and is still considered a great challenge even for state-of-the-art deep learning models \cite{testolin2023can}. The goal of the system is to generate the final solution of arithmetical expressions that can involve a variable number of nested elementary operations (sums, subtractions, and multiplications) delimited by parentheses. 
The model is trained on a limited subset of possible problems that contain only up to 2 nested operations. In order to evaluate its extrapolation capabilities, in the testing phase the model is probed with problems involving up to 10 nested operations.

Our results show that the proposed hybrid architecture achieves better out-of-distribution generalization compared to a Transformer encoder-decoder trained to solve the same problem end-to-end, and to a large language model tested on the same task, adapting the original inputs to create appropriate prompts\footnote{Our code is available at \url{https://github.com/flavio2018/nesy23}.}.

%% file: 1.5-system.tex
\begin{figure}
    \centering
\includegraphics[width=\textwidth]{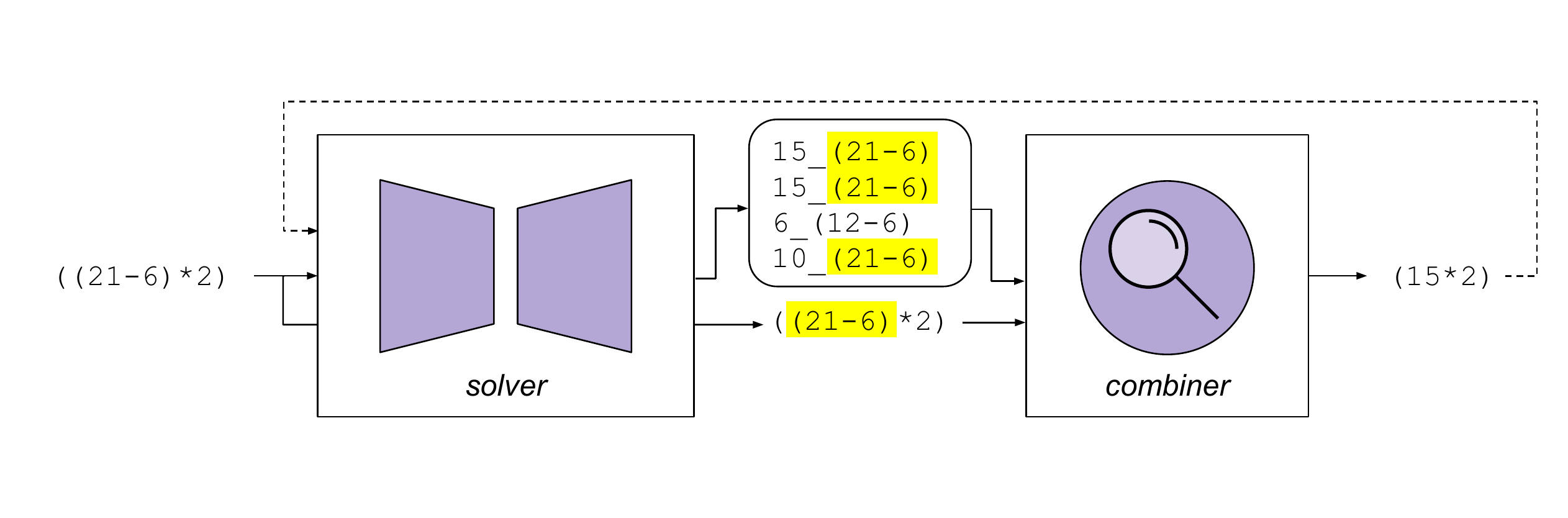}
    \caption{A visual representation of the hybrid system. The system takes an arithmetical expression as input. The solver learns to generate a set of candidate simplifications, which are produced as output according to a predefined syntax that describes the result of the operation concatenated with the target substitution. The combiner then filters well-formed outputs, selects the most frequently generated candidate and applies the corresponding replacement of the target sub-string. The output of the combiner is then fed back to the system, enabling the iterative solution of complex problems.\label{fig:system}} 
\end{figure}

The proposed architecture is composed of two parts (Figure~\ref{fig:system}): a trainable seq2seq model which we call the `solver', and a deterministic module named the `combiner'.
The system operates by pipelining the computation of the solver and the combiner, and then applying the pipeline iteratively.

We can define $E$ as the set of nested arithmetical expressions represented as strings of symbols on which we test our model, and $R \subset E$ as the subset of expressions that consist only of integer values, and thus represent the result (or values) of complex expressions.
Furthermore, since each $e \in E \setminus R$ is composed of nested expressions, for each $e$ we can identify one innermost sub-expression $s_k$ (only one in our specific case, see Data in Section \ref{par:methods}) which can be solved independently from the context (we can define $r_k \in R$ as the result of $s_k$). Finally, we can define a function which associates to each $e \in E$ a chain of solution steps: $chain(e) = (e_k, e_{k-1}, ..., e_0)$, where $e, e_j \in E,\ j\in\{1,\ldots,k\}$, $e_0 \in R$ is the final solution of $e_k$, and $e_{j-1}$ can be described as $e_j$ where the innermost sub-expression $s_j$ has been solved.

Given the compositional nature of the problem, we can distinguish between the goal of solving the entire expression $e$, i.e. finding $e_0$, and the sub-goals of reducing sub-expressions, i.e. computing the result $r_j$ of each sub-expression $s_j,\ j\in\{1,\ldots,k\}$.

 In our approach, the solver is trained to match arithmetical sub-expressions in their context to the corresponding numerical solution, by generating as output a string composed of two parts: a replacement candidate (i.e. the result of the innermost sub-expression $r_j$) and a replacement target (i.e. the innermost sub-expression itself $s_j$), concatenated using the `underscore' symbol.
For example, if the model receives as input the expression \texttt{((21-6)*2)} it should produce as output the string \texttt{15\_(21-6)}.
More formally, the solver can be defined as a function $solver\colon E \rightarrow R \times E$, $solver(e_j) = (r_j, s_j)$ where $e_j\in E$ is the nested input expression, $s_j \in E$ is the innermost sub-expression in $e_j$, and $r_j \in R$ is the result of $s_j$.

The combiner receives the same string $e_j$ provided in input to the solver together with the solver's output, i.e. $(r_j,s_j)$, and produces $e_{j-1}$, i.e. the simplified version of $e_j$ where the sub-expression $s_j$ has been substituted by $r_j$. $e_{j-1}$ can be fed back into the system until the problem is completely solved (note that the combiner module is not required during training of the seq2seq model).
Formally, the combiner can be defined as a function \mbox{$combiner\colon E\times (R \times E) \rightarrow E, \ combiner(e_j, (r_j,s_j)) = e_{j-1}$}.

Finally, the whole hybrid system can be defined as a function $hybrid\colon E \rightarrow E$, \mbox{$hybrid(e_j) = combiner(e_j, solver(e_j))= e_{j-1}$}.
Since the output domain of $hybrid$ is the same as its input domain, it can be applied iteratively on its own output to solve arithmetical expressions step by step. Indeed, the system receives the expression represented as a sequence of characters as input and outputs a reduced version of the expression, which is fed back in the system unless the output only contains a positive or negative integer.

\subsection{The solver}
The solver is a standard Transformer encoder-decoder \citep{Vaswani2017AttentionIA} with a one-layer encoder and a one-layer decoder.
Our version of the model differs from the standard Transformer only due to the use of label positional encodings \citep{li2022systematic} rather than standard sinusoidal positional encodings.
Label positional encodings have recently been proposed as a method to improve out-of-distribution generalization of a Transformer decoder in simple algorithmic tasks \citep{li2022systematic}.
In this work, we have applied this technique to an encoder-decoder Transformer exploring the extent to which these encodings can be applied to the solution of arithmetic problems, a task that is relatively similar but slightly more complex than simple algorithmic tasks.
Given a sequence of $k$ symbols, the corresponding sinusoidal positional encodings \citep{Vaswani2017AttentionIA} are the first $k$ vectors in a sequence of $m, m \geq k$ vectors that carry a position signal which is generated according to the formulas $PE_{(pos,2i)} = sin(pos / 10000^{2i/d_{model}}), \ \  PE_{(pos,2i+1)} = cos(pos / 10000^{2i/d_{model}})$,
where $pos$ is the position of the token in the sequence and $i$ is the dimension of the vector.
When using label positional encodings, to associate a position signal to a sequence of $k$ symbols, rather than considering the first $k$ sinusoidal encodings, we generate the position signal sampling $k$ integers in the interval $[0, m-1]$, sorting them in ascending order and then taking the corresponding sinusoidal positional encodings.

\subsection{The combiner}
The purpose of the combiner is to deterministically replace part of the input string with the candidate solution proposed by the solver model. To enable the deterministic substitution in the combiner, the output of the solver must contain the expression and its numerical solution formatted according to a specific syntax. However, this may not be the case: we call `halted sequences', the ones for which the solver output is not compliant with the required format. Indeed, in such cases the processing of the sequence should stop because there is no meaningful way to proceed with the simplification.

In order to reduce the number of halted sequences, we have chosen to employ a multi-output generation strategy.
For each input expression, we generate $N$ outputs sampling from the distributions obtained from the softmaxed logits returned by the solver decoder. Since the Transformer generates the output in an auto-regressive fashion, for each token in the output sequence we sample from the distribution and feed back the updated sequence of output tokens to the decoder.
The set of $N$ outputs is then given as input to the combiner, which first filters well-formed outputs, then selects the most frequently generated sequence, and finally applies the substitution of the target with the candidate in the input expression.

%% file: 2-method.tex
{\noindent \bf Data}
One of the goals in the design of our hybrid system is to combine the generalization capabilities of the solver with the deterministic nature of the combiner, which should allow learning by using a relatively small subset of the data. Intuitively, this subset should include the minimal amount of examples necessary for the model to learn the base solution step of a complex expression.
We thus include in the training set only expressions with up to two nested operations between integers of up to two digits.

For simplicity, in all dataset splits and nesting levels, we build expressions that have only one operation for each nesting level. Therefore, the model never faces the ambiguous situation in which there are multiple operations that could be solved at the same time.

Since at test time the full hybrid system will be evaluated on expressions with up to $10$ nested operations, it may happen that test expressions have intermediate or final results with more than two digits.
We therefore take some measures to guarantee that the system will behave with more complex expressions.
In particular, to ensure that the distribution of the expressions that will be used to test the model is coherent with the one that was used to train it, we limit both training and testing expressions to the ones that have intermediate and final results with at most two digits.
Furthermore, we ensure that at training time the Transformer observes examples of a complete solution path of an arithmetical expression, i.e. all the steps leading to the final result.
Hence, we ensure that the first-level simplification of some training samples involving two nested operations is also present in the training set.

To test out-of-distribution generalization more easily, we sample batches of data on-the-fly from the distribution described above, parameterized by the number of operations and size of operands in the expression.
Given the data generation procedure and the number of expressions that could virtually be generated, the actual number of training examples observed by the model depends on the number of learning iterations. 
To ensure the separation of the training, validation and test splits, we use both a pre-computed split that ensures some of the training samples have a full solution path, as well as a hash-based split that sorts all remaining samples in three different splits on-the-fly.

{\noindent \bf Evaluation Metrics}
We use two metrics in the performance evaluation of all systems: character and sequence accuracy.
Given a batch of target sequences, we define character accuracy as the percentage of correct characters in the output of a system, and sequence accuracy as the percentage of output sequences that match the target exactly.
When measuring the performance of the solver and that of the hybrid system, these two metrics are applied to different kinds of output, as the solver and the hybrid system solve two different tasks.
Both metrics, in the case of the hybrid system, are computed considering halted sequences as mistakes, i.e. the output sequence is assumed to be completely different from the target.

{\noindent \bf Model training}
The Transformer is trained in an auto-regressive regime, i.e. without teacher forcing.
We have trained all models for 150k updates using Adam optimizer with a learning rate of $1e-4$ and batch size of 128.\\
To choose the best configuration for the solver hyperparameters we performed a random search on the number of heads and the dimensionality of embeddings.
For both the Transformer encoder and decoder we have compared the performance of models with 4 and 8 heads, 256, 512 and 1024-dimensional embeddings, and 256, 512 and 1024-dimensional hidden feed-forward layer on the in-distribution validation set.
We finally selected the solver model with 8 attention heads, a 1024-dimensional state and a 1024-dimensional feed-forward layer.

Using the multi-output generation strategy, the number of outputs generated for each input is an hyperparameter whose choice clearly requires trading model accuracy for speed.
We compared the performance of the hybrid system on the solution of arithmetical expressions using $N=1, 20, 40, 60, 80, 100$ generated outputs per input.
We have observed that sampling as few as 20 outputs for each input already guarantees much better results than a 1-shot output generation, especially in terms of surviving sequences.
Nevertheless, we chose to generate 100 outputs as we observed marginal performance gains when continuing to increase the number of outputs.

%% file: 3-results.tex
\begin{table}
    \centering
    \resizebox{\textwidth}{!}{%
    \input{tables/model-size-comparison}}
    \caption{Character (top rows) and sequence (bottom rows) accuracy of three different solver models on the sub-expression solution task. The models have 256-, 512- and 1024-dimensional embeddings, respectively. We report average and standard deviation of both metrics over 10 batches of 100 sequences. The figures show that the label positional embeddings endow the solver with strong out-of-distribution generalization capability on the sub-expression solution task.\label{tab:models_comparison}} 
\end{table}

Given the hybrid nature of our system, in order to measure its performance it is meaningful to consider both the trained solver module in isolation, and the hybrid system as a whole.

We also compare the performance of the hybrid system with that of a Transformer encoder-decoder trained to directly output the result of arithmetical expressions.
Finally, we compare the performance of the system with that of a Large Language Model fine-tuned for text completion.

\subsection{Solver}
We hypothesize that the solver module can leverage label positional encodings to achieve out-of-distribution generalization capacity on the sub-expression solution task.
Therefore, the solver receives at test time complex arithmetical expressions with up to 10 nested operations, i.e. up to 8 more nested operations than seen during training.
The goal of the model remains to output the result of the innermost expression contained in the input (replacement candidate) concatenated to the expression itself (replacement target).

\begin{figure}
\centering
\begin{subfigure}{.47\textwidth}
  \centering
  \includegraphics[width=\linewidth]{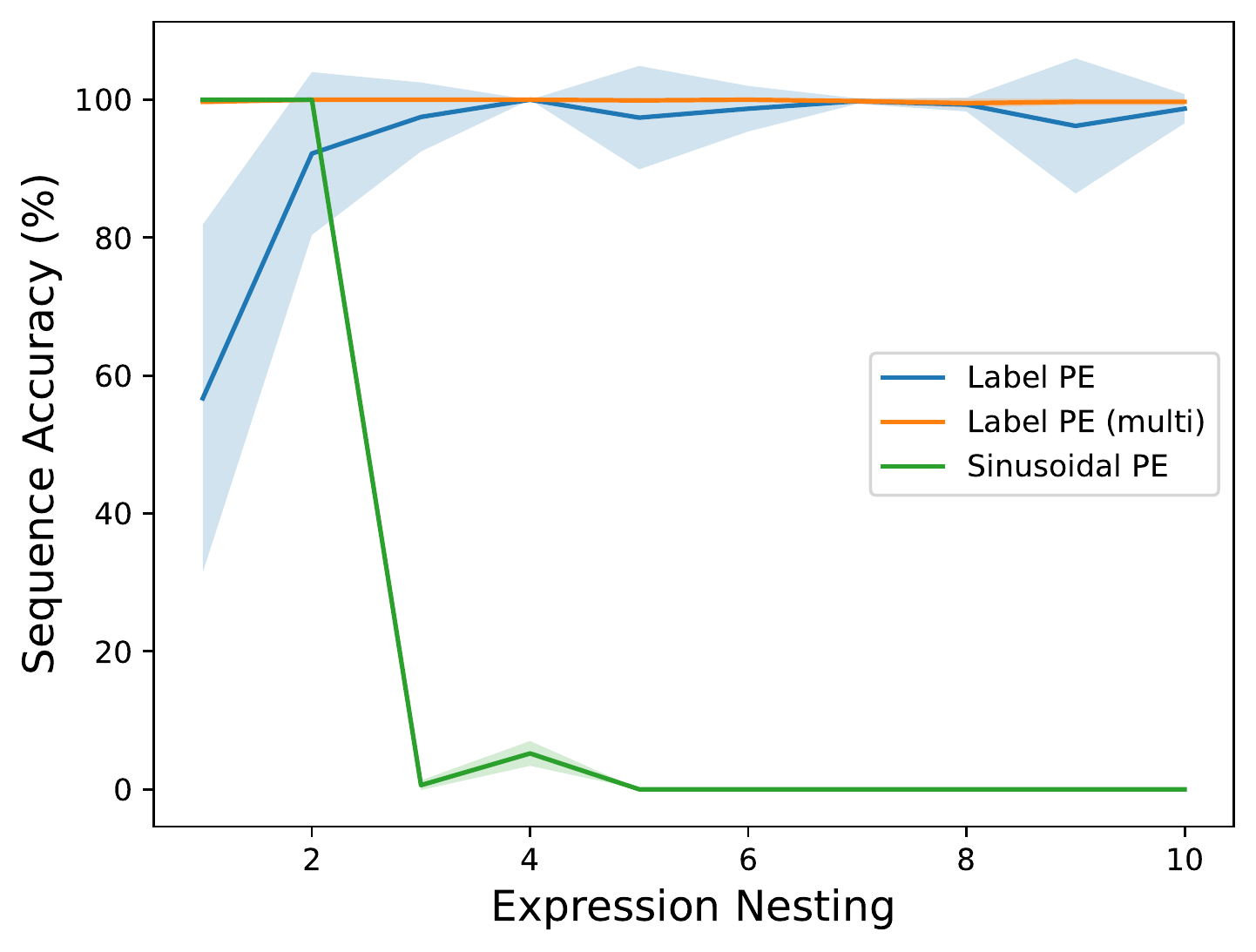}
  \caption{Performance of solver modules trained with Label and Sinusoidal Positional Encodings. \texttt{Label PE (multi)} refers to a solver that generates 100 outputs per input, applying the same filtering rules as the combiner.}
  \label{fig:lpe_vs_nolpe}
\end{subfigure}%
\begin{subfigure}{.47\textwidth}
  \centering
  \includegraphics[width=\linewidth]{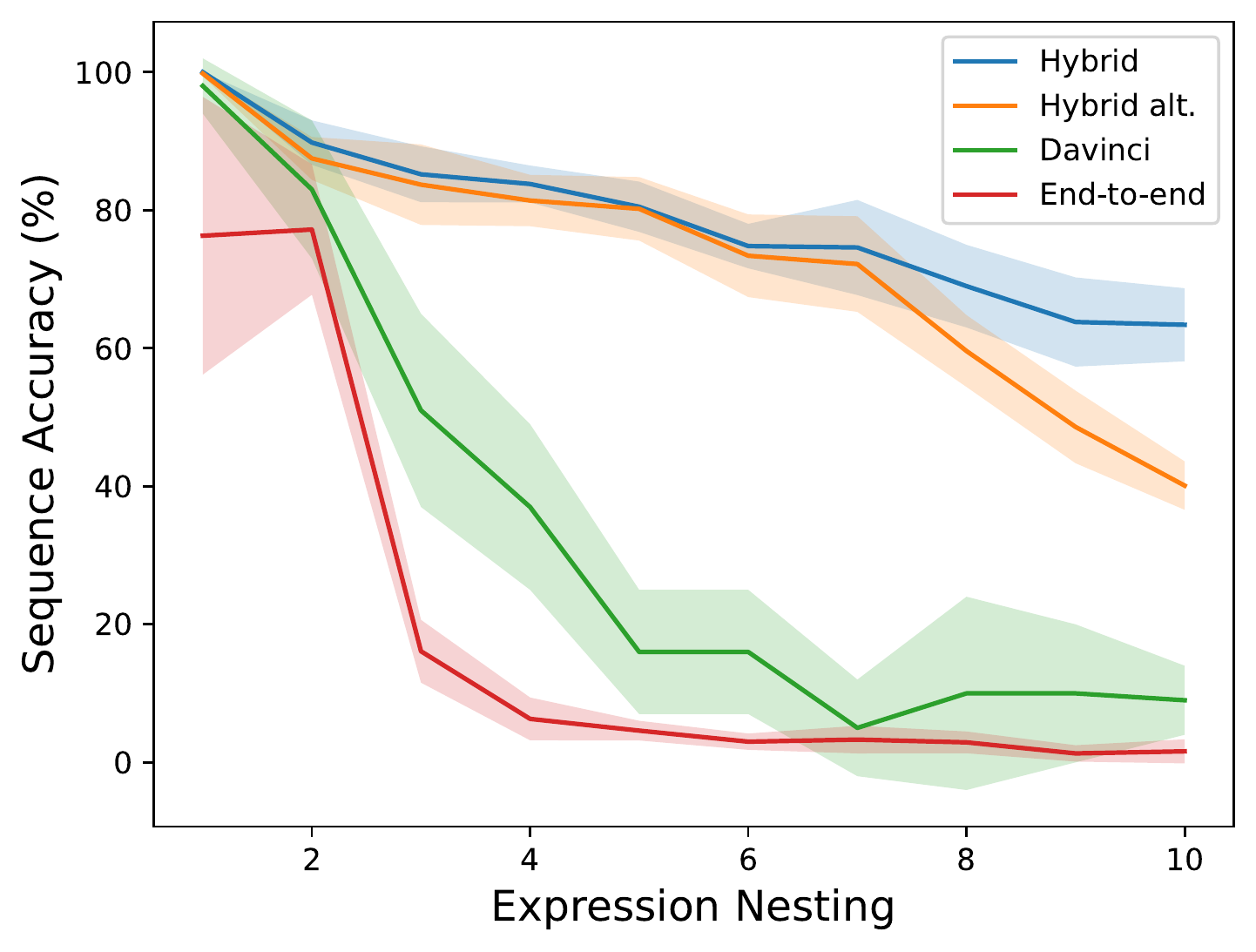}
  \caption{Performance of hybrid system, end-to-end and LLM baselines. \texttt{Hybrid alt.} refers to a hybrid system where the combiner selects the most frequent outputs before selecting well-formed outputs.}
  \label{fig:main_figure}
\end{subfigure}
\caption{Sequence accuracy of all models we consider. The solver modules (left) are tested on the sub-expression solution task, while the hybrid systems and baseline models (right) are tested on the task of solution of the full expression. In all cases we report mean and standard deviation computed across 10 batches of 100 sequences per nesting value, except for the Davinci model for which batches contain 10 sequences.}
\label{fig:test}
\end{figure}

The sequence accuracy of the solver on the sub-expression solution task is represented in Figure \ref{fig:lpe_vs_nolpe}, which shows that the use of Label Positional Encodings enables out-of-distribution generalization. 
We also report in Table \ref{tab:models_comparison} the performance of solver models with different embedding dimensionality to provide further insight on the impact of label positional encodings.
Indeed, our results show that, independently of model size, a Transformer encoder-decoder using label positional encodings can generalize to out-of-distribution samples on this task.

Figure \ref{fig:lpe_vs_nolpe} shows that using Label Positional Encodings also leads to a lower performance on test expressions, especially on the nesting level 1.
However, this has no impact on the functioning of the hybrid system when the multi-output generation strategy is used.

\subsection{Hybrid System}
We then test the entire hybrid system on the arithmetic expressions resolution task. 

Figure \ref{fig:main_figure} shows the sequence accuracy of the hybrid system exploiting the multi-output generation strategy with 100 outputs per input on arithmetical expressions involving up to 10 nested operations.
The performance metric shows that the combination of the architectural elements in our model allows to produce exact answers to harder problems than seen during training in a large fraction of cases.
The slow performance decline is proportional to expression complexity, as it is due to the accumulation of errors committed by the solver module in several iterations on the same expression.
We have also experimented on a possible alternative combiner mechanism, where the selection of the most frequently generated output for a given input is done before filtering ill-formed outputs.
Figure \ref{fig:main_figure} shows that the method (Hybrid alt.) globally still achieves superior performance compared to baselines, but has worse performance with respect to the default combiner mechanism in harder problems due to the higher percentage of halted sequences, as shown in Table \ref{tab:hybrid_no_filter}.
The similar sequence accuracy of the two methods on expressions with fewer nested operations is due to the fact that the default combiner mechanism, despite halting fewer sequences, commits more semantic errors on the final results.

In the Appendix we provide further experimental evidence to shed light on the functioning of the proposed system.
For completeness, we report character accuracy of the models we consider on the expressions resolution task in Table \ref{tab:performance_comparison}.
In Table \ref{tab:n_samples_comparison} we provide further insight on the impact of the number of outputs per input on both the accuracy of the hybrid system and the percentage of halted sequences.
Finally, in Table \ref{tab:emb_dim_comparison}  we report out-of-distribution performance metrics for hybrid systems built with differently-sized solvers to provide experimental evidence on the impact of model size on the generalization capacity of the system.

\begin{table}[]
    \centering
    \resizebox{\textwidth}{!}{%
    \input{tables/filter_nofilter_comparison}}
    \caption{Percentage of halted sequences using two different combiner mechanisms. The top row is relative to the default combiner mechanism, while the bottom row refers to a combiner where the most frequently generated outputs are selected before filtering out ill-formed ones.}
    \label{tab:hybrid_no_filter}
\end{table}

\subsection{Comparison with end-to-end model}
To motivate empirically the presence of a deterministic component in our hybrid system, we compare the results of our system on the solution of arithmetical expressions with those obtained on the same problem by a Transformer trained to directly output the result of a given arithmetical expression.
Also in this case, we use label positional encodings on the Transformer encoder and decoder inputs, in order to control the effect of this architectural element on the performance of the system.
The training hyperparameters and dataset splits are the same used to train the seq2seq part of the hybrid system.
Furthermore, the Transformer has the same architecture and model size as the trained component in our system.

Figure \ref{fig:main_figure} represents the sequence accuracy of the end-to-end trained system on this task.
It is evident that, while on in-distribution samples the end-to-end trained Transformer has reasonably good performance, it nevertheless fails to generalize the learned solution process on longer and more complex expressions.
\subsection{Comparison with a Large Language Model}
Finally, we compare our hybrid system with a Large Language Model (LLM), since these models often achieve state-of-the-art performance in sophisticated reasoning tasks \citep{testolin2023can}.
We consider the outputs provided by a version of GPT-3 \citep{NEURIPS2020_1457c0d6}, more precisely a variant of the model iteration generally referred to as GPT-3.5.
The version of the model we tested is the one referred to as \texttt{text-davinci-003} in the OpenAI API at the time of writing.
According to OpenAI's official website \footnote{\url{https://platform.openai.com/docs/model-index-for-researchers}}, the model is an InstructGPT model \citep{ouyang2022training} fine-tuned for text completion.

We prompt the model with an adapted version of our inputs, first showing the model an example of a correctly solved expression and then asking to complete the solution of a new expression.
For example, given the expression \texttt{(((3*2)-2)+5)}, a possible corresponding prompt could be \texttt{((2+4)*6)=36<END>\textbackslash n(((3*2)-2)+5)=}.
We then require via API to stop the generation of the output string when the model outputs the \texttt{<END>} token.
In this way, we ensure that the vast majority of the model outputs is the solution of the input expression proposed by the model.

Figure \ref{fig:main_figure} shows the performance of the model in solving arithmetic expressions.
Despite the gigantic size of the model and the amount of training data, the model cannot directly output the solution of complex arithmetical expressions.

%% file: tables/model-size-comparison.tex
\setlength{\tabcolsep}{3pt}
\begin{tabular}{llrrrrrrrrrr}
\toprule
&Nesting &    1  &    2  &    3  &    4  &    5  &    6  &    7  &    8  &   9  &    10 \\
\midrule
\multirow{3}{*}{\rotatebox[origin=c]{90}{Char Acc}}
&small &  57.2±17.5 &  92.1±15.2 &  99.2±1.3 &   97.6±4.9 &  98.4±4.5 &  99.5±1.2 &  99.9±0.2 &  99.3±0.9 &  98.1±2.8 &  98.9±1.6 \\
&medium     &  72.2±17.4 &   96.8±5.4 &  99.7±1.0 &  100.0±0.0 &  100.0±0.0 &  99.7±0.8 &  100.0±0.0 &  100.0±0.1 &  99.5±1.5 &  99.6±1.2 \\
&large     &  78.5±12.6 &   96.7±4.6 &  98.8±2.8 &  100.0±0.0 &  98.5±4.5 &  99.5±1.6 &  99.9±0.2 &  99.9±0.1 &  97.6±6.2 &  99.5±0.8 \\
\midrule
\multirow{3}{*}{\rotatebox[origin=c]{90}{Seq Acc}}
&small      &  28.1±25.4 &  82.3±27.8 &  98.5±2.6 &  93.5±12.9 &  97.1±8.0 &  99.2±2.1 &  99.4±1.5 &  98.4±1.9 &  96.2±6.1 &  97.4±3.5 \\
&medium      &  50.8±27.4 &  93.6±11.0 &  99.1±2.7 &   99.9±0.3 &  100.0±0.0 &  99.2±1.8 &  100.0±0.0 &   99.8±0.6 &  99.3±1.6 &  99.2±2.4 \\
&large      &  56.7±25.2 &  92.2±11.8 &  97.5±5.0 &  100.0±0.0 &  97.4±7.5 &  98.7±3.3 &  99.8±0.4 &  99.3±1.0 &  96.2±9.8 &  98.7±2.1 \\
\bottomrule
\end{tabular}

%% file: tables/filter_nofilter_comparison.tex
\begin{tabular}{lrrrrrrrrrr}
\toprule
Nesting &        1  &        2  &        3  &        4  &        5  &        6  &        7  &        8  &        9  &        10 \\

\midrule
Hybrid  &    0.0±0.0 &   0.0±0.0 &   0.0±0.0 &   0.0±0.0 &   0.1±0.3 &   0.3±0.5 &   0.2±0.4 &   0.8±0.7 &   1.4±0.9 &   2.1±1.4 \\
Hybrid alt. & 0.1±0.3 &   5.1±1.8 &   6.1±2.1 &   5.8±3.2 &   8.3±3.3 &  10.1±3.9 &  12.2±5.0 &  25.6±3.9 &  36.0±5.9 &  47.7±5.3 \\
\bottomrule
\end{tabular}

%% file: 4-related.tex
In this work, we explored the idea that neural networks can learn to solve problems that require composition of elementary solution steps, like arithmetic expressions, by iteratively applying a learned substitution rule on the input string.
Other works have recently explored similar ideas, often in different kind of problems.
In \citep{Deshpande2021RecTAR} the authors propose a Transformer model that is trained to iteratively solve arithmetical expressions involving sums and subtractions.
Differently from our system, it requires twice the number of iterations as it is trained to first highlight and then solve relevant parts of the input expressions.
The authors of \citep{DBLP:journals/corr/abs-2201-11766} propose a Transformer-based architecture to iteratively solve compositional generalization tasks.
In this work, authors exploit relative positional encodings \citep{shaw-etal-2018-self} to enable out-of-distribution generalization.
Differently from our case, the proposed architecture includes an ad-hoc trainable component named copy-decoder, designed to facilitate learning to copy parts of the input in the output.
The authors of \citep{NEURIPS2018_0e64a7b0} propose two neural architectures that incorporate ad-hoc learning biases to learn addition and multiplication generalizing beyond the training range.
However, their approach is limited to arithmetic problems and cannot be extended to more general symbolic reasoning problems.
In \citep{DBLP:journals/corr/abs-2110-04169} the idea of iteratively processing the input to learn a systematic solution strategy is applied to the problem of visual navigation with a multimodal architecture.

Recently, researchers have also focused on studying whether standard seq2seq models can generalize to out-of-distribution samples in compositional tasks.
In \citep{csordas-etal-2021-devil} the authors show that relative positional encodings, carefully tuned early stopping and scaling of encodings can improve the performance of a Universal Transformer \citep{dehghani2018universal} on compositional generalization tasks.
In a follow-up work \citep{csordas2022ndr}, the same authors propose an approach to the solution of algorithmic tasks similar to the one we propose in this work.
Their proposal consists of learning step-by-step solution strategies modifying the architecture of a Transformer encoder.
Differently from our case, the authors propose to recursively modify internal representations, rather than iteratively manipulating symbolic, external representations.

Another recent approach to exploit deep learning models to solve reasoning problems is to embed neural networks into existing automatic reasoning frameworks, such as logic or probabilistic programming languages \citep{ijcai2021p254, NEURIPS2018_dc5d637e}. In such frameworks, systematic generalization is achieved by leveraging constructs in the pre-existing programming languages, supporting theoretical standpoints arguing that human cognition should be understood as a hybrid of neural and algorithmic mechanisms \citep{Marcus2001}.
In our work, instead, we are interested in exploring the possibility that the solution process of symbolic reasoning problems can be implemented with neural systems, leveraging simpler additional mechanisms than the ones used in the works mentioned above, and by injecting learning biases in the neural system to allow the iterative solution of symbolic reasoning problems exploiting their recursive structure.

%% file: 5-discussion.tex
In this paper we have proposed a hybrid system that leverages the combination of a trained module and a deterministic one to achieve strong generalization abilities on the solution of simple arithmetical expressions.
While we have only considered this specific task, it should be emphasized that the framework described in section \ref{par:system} can be used to describe a more general class of symbolic problems where the solution can be derived iteratively.
Indeed, one can easily imagine cases (such as symbolic mathematics) in which starting from the initial symbolic expression one can derive a tree - rather than a chain - of syntactically correct expressions.
Furthermore, each expression could be the result of a more generic transformation than a solution or simplification step, which may even make the expression longer.
As a consequence, the hybrid system we propose could in principle be adapted to a wider range of symbolic problems than the one considered here.
In future work, we plan to explore this aspect, experimenting on tasks like symbolic mathematics, algorithmic tasks and program synthesis.

%% file: 6-appendix.tex
\begin{table}[h]
    \centering
    \resizebox{\textwidth}{!}{%
    \input{tables/performance_comparison}}
    \caption{Character accuracy of the three models. We report mean and standard deviation computed across 10 batches of 100 sequences per nesting value, except for the Davinci model for which batches contain 10 sequences.\label{tab:performance_comparison}}
\end{table}

\begin{table}[h]
    \centering
    \resizebox{\textwidth}{!}{%
    \input{tables/n_samples_comparison}}
    \caption{Comparison of hybrid model performance varying the number of samples generated using the multi-output generation strategy. Figures show that increasing the number of generated samples strongly impacts the percentage of halted sequences, especially for more complex expressions.}
    \label{tab:n_samples_comparison}
\end{table}

\begin{table}[h]
    \centering
    \resizebox{\textwidth}{!}{%
    \input{tables/emb_dim_comparison}}
    \caption{Comparison of hybrid model performance varying the embedding dimensionality of the solver, using a multi-output generation strategy with 100 outputs per sequences. Smaller models make more mistakes and lead both to higher percentage of halted sequences and to lower accuracy of the final results.}
    \label{tab:emb_dim_comparison}
\end{table}

%% file: tables/performance_comparison.tex
\setlength{\tabcolsep}{3pt}
\begin{tabular}{lrrrrrrrrrr}
\toprule
Nesting &         1  &         2  &         3  &         4  &         5  &         6  &         7  &         8  &         9  &         10 \\
\midrule

Transformer &  89.1±10.3 &   89.7±4.6 &   56.0±3.3 &   47.4±2.5 &   44.9±2.2 &   43.3±1.9 &   42.2±2.1 &   40.8±2.6 &   40.3±1.9 &   39.7±1.1 \\
Davinci &  98.0±5.0 &   89.0±8.0 &  69.0±10.0 &  59.0±10.0 &  41.0±10.0 &  39.0±12.0 &  34.0±8.0 &  35.0±14.0 &   29.0±9.0 &  31.0±9.0 \\
Hybrid &  100.0±0.0 &  93.8±1.9 &  92.2±2.7 &  91.4±2.5 &  89.8±2.9 &  86.0±2.9 &  85.1±4.7 &  81.6±4.0 &  77.0±4.8 &  74.1±3.6 \\
Hybrid alt. &  99.9±0.2 &  91.0±2.7 &  89.4±3.9 &  88.3±3.5 &  86.9±3.2 &  83.0±4.3 &  81.4±5.0 &  69.0±4.5 &  61.4±4.6 &  53.0±3.1 \\


\bottomrule
\end{tabular}

%% file: tables/n_samples_comparison.tex
\setlength{\tabcolsep}{3pt}
\begin{tabular}{llrrrrrrrrrr}
\toprule
& Nesting &         1  &        2  &        3  &        4  &        5  &        6  &        7  &        8  &        9  &        10 \\
\midrule

\multirow{6}{*}{\rotatebox[origin=c]{90}{Char Acc}}
& 1-shot   &  59.7±21.5 &  44.6±20.4 &  52.1±20.1 &  42.4±19.9 &  33.3±15.6 &  28.8±15.3 &  26.2±10.2 &   24.9±9.8 &  22.5±4.9 &  22.0±10.8 \\
& 20 samples &  99.5±0.8 &  92.4±2.3 &  89.4±4.1 &  88.4±3.6 &  88.0±4.7 &  85.6±3.9 &  80.8±4.9 &  76.6±3.5 &  72.0±6.0 &  67.3±3.4 \\
& 40 samples &  99.8±0.4 &  92.4±2.8 &  91.6±2.2 &  90.9±2.9 &  88.0±2.8 &  87.0±3.1 &  83.7±3.4 &  79.6±3.6 &  74.9±4.9 &  70.0±5.0 \\
& 60 samples &  100.0±0.1 &  92.6±2.2 &  90.4±3.2 &  88.5±4.8 &  89.9±3.4 &  88.3±2.6 &  85.1±2.6 &  83.8±3.7 &  76.9±2.9 &  72.6±5.6 \\
& 80 samples &  99.9±0.3 &  94.9±1.3 &  91.4±2.4 &  90.8±2.2 &  90.2±2.6 &  89.0±3.7 &  85.5±1.4 &  80.6±3.5 &  79.5±3.9 &  70.8±3.5 \\
& 100 samples &  100.0±0.0 &  93.8±1.9 &  92.2±2.7 &  91.4±2.5 &  89.8±2.9 &  86.0±2.9 &  85.1±4.7 &  81.6±4.0 &  77.0±4.8 &  74.1±3.6 \\

\midrule

\multirow{6}{*}{\rotatebox[origin=c]{90}{Seq Acc}}
& 1-shot  &  48.1±26.1 &  32.9±24.6 &  41.9±22.3 &  33.7±19.6 &  24.0±16.0 &  18.5±15.0 &  13.7±12.1 &  11.2±10.9 &   9.9±6.1 &    8.9±7.8 \\
& 20 samples  &  98.6±2.1 &  86.8±4.2 &  82.0±5.6 &  78.9±7.2 &  78.3±7.2 &  75.0±6.7 &  66.3±5.8 &  64.3±7.3 &  58.1±7.1 &  55.2±5.5 \\
& 40 samples  &  99.6±0.8 &  86.7±4.6 &  84.7±3.7 &  84.2±4.4 &  78.5±3.8 &  76.9±4.8 &  72.3±4.8 &  65.8±7.4 &  63.6±7.0 &  58.6±6.8 \\
& 60 samples  &   99.9±0.3 &  87.6±4.1 &  82.8±4.3 &  80.5±5.3 &  81.5±5.3 &  77.9±4.7 &  74.7±4.1 &  73.6±6.5 &  64.9±5.8 &  62.0±5.9 \\
& 80 samples  &  99.6±0.9 &  89.7±1.8 &  84.7±3.9 &  83.1±3.4 &  81.8±5.6 &  78.8±5.4 &  74.5±3.3 &  69.2±4.4 &  67.3±5.0 &  58.9±6.3 \\
& 100 samples  &  100.0±0.0 &  89.8±3.2 &  85.2±4.0 &  83.8±2.7 &  80.5±3.6 &  74.8±3.2 &  74.6±6.9 &  69.0±6.0 &  63.8±6.5 &  63.4±5.3 \\

\midrule

\multirow{6}{*}{\rotatebox[origin=c]{90}{Halted}}
& 1-shot      &  47.5±23.8 &  59.3±27.5 &  51.2±24.6 &  50.4±24.5 &  58.9±24.6 &  66.6±27.5 &  81.7±15.3 &  84.5±14.0 &  85.9±8.1 &  76.5±27.2 \\
& 20 samples      &   0.0±0.0 &   0.8±0.7 &   0.7±1.3 &   2.3±2.1 &   1.9±0.8 &   1.7±1.6 &   2.2±1.9 &   5.1±2.1 &   7.4±4.0 &  11.7±3.3 \\
& 40 samples      &   0.0±0.0 &   0.1±0.3 &   0.1±0.3 &   0.1±0.3 &   0.5±0.5 &   0.2±0.4 &   1.0±1.0 &   2.1±1.1 &   4.5±1.8 &   7.6±2.3 \\
& 60 samples      &    0.0±0.0 &   0.0±0.0 &   0.1±0.3 &   0.0±0.0 &   0.2±0.4 &   0.1±0.3 &   0.3±0.5 &   1.9±1.4 &   2.5±1.4 &   4.3±2.0 \\
& 80 samples      &   0.0±0.0 &   0.0±0.0 &   0.1±0.3 &   0.0±0.0 &   0.0±0.0 &   0.1±0.3 &   0.3±0.5 &   0.7±0.9 &   2.3±1.3 &   4.1±2.5 \\
& 100 samples      &    0.0±0.0 &   0.0±0.0 &   0.0±0.0 &   0.0±0.0 &   0.1±0.3 &   0.3±0.5 &   0.2±0.4 &   0.8±0.7 &   1.4±0.9 &   2.1±1.4 \\
\bottomrule
\end{tabular}

%% file: tables/emb_dim_comparison.tex
\setlength{\tabcolsep}{3pt}
\begin{tabular}{llrrrrrrrrrr}
\toprule
& Nesting &         1  &        2  &        3  &        4  &        5  &        6  &        7  &        8  &        9  &        10 \\
\midrule

\multirow{3}{*}{\rotatebox[origin=c]{90}{Char Acc}}

& small &  98.8±1.1 &  92.9±2.0 &  87.3±2.2 &  83.6±2.2 &  79.4±4.2 &  70.6±2.6 &  61.1±3.6 &  50.7±4.6 &  46.2±4.2 &  39.4±3.4 \\
& medium &  100.0±0.1 &  93.9±1.4 &  91.9±1.6 &  90.9±2.5 &  90.0±1.6 &  86.4±2.1 &  78.7±3.7 &  72.1±3.9 &  66.4±5.2 &  62.2±3.6 \\
& large &  100.0±0.0 &  93.8±1.9 &  92.2±2.7 &  91.4±2.5 &  89.8±2.9 &  86.0±2.9 &  85.1±4.7 &  81.6±4.0 &  77.0±4.8 &  74.1±3.6 \\
\midrule

\multirow{3}{*}{\rotatebox[origin=c]{90}{Seq Acc}}
& small  &  97.3±2.4 &  84.2±4.0 &  74.8±4.0 &  66.5±4.8 &  61.8±6.0 &  50.2±5.7 &  45.8±4.2 &  33.4±5.2 &  30.1±2.9 &  23.0±4.1 \\
& medium  &   99.9±0.3 &  85.6±2.4 &  80.6±2.5 &  77.7±4.7 &  74.6±3.7 &  69.9±5.9 &  60.2±5.2 &  55.6±6.2 &  50.1±7.1 &  44.8±6.1 \\
& large  &  100.0±0.0 &  89.8±3.2 &  85.2±4.0 &  83.8±2.7 &  80.5±3.6 &  74.8±3.2 &  74.6±6.9 &  69.0±6.0 &  63.8±6.5 &  63.4±5.3 \\

\midrule

\multirow{3}{*}{\rotatebox[origin=c]{90}{Halted}}
& small      &   0.0±0.0 &   0.5±0.8 &   1.5±0.7 &   1.3±0.9 &   2.3±1.6 &   7.7±2.6 &  18.5±4.5 &  27.3±3.7 &  39.4±4.9 &  51.9±5.6 \\
& medium      &    0.0±0.0 &   0.0±0.0 &   0.1±0.3 &   0.1±0.3 &   0.0±0.0 &   0.1±0.3 &   1.2±0.9 &   4.5±2.1 &  11.5±4.0 &  17.5±2.9 \\
& large      &    0.0±0.0 &   0.0±0.0 &   0.0±0.0 &   0.0±0.0 &   0.1±0.3 &   0.3±0.5 &   0.2±0.4 &   0.8±0.7 &   1.4±0.9 &   2.1±1.4 \\
\bottomrule
\end{tabular}

%% file: main.bbl
\begin{thebibliography}{19}
\expandafter\ifx\csname natexlab\endcsname\relax\def\natexlab#1{#1}\fi
\providecommand{\url}[1]{\texttt{#1}}
\providecommand{\href}[2]{#2}
\providecommand{\path}[1]{#1}
\providecommand{\DOIprefix}{doi:}
\providecommand{\ArXivprefix}{arXiv:}
\providecommand{\URLprefix}{URL: }
\providecommand{\Pubmedprefix}{pmid:}
\providecommand{\doi}[1]{\href{http://dx.doi.org/#1}{\path{#1}}}
\providecommand{\Pubmed}[1]{\href{pmid:#1}{\path{#1}}}
\providecommand{\bibinfo}[2]{#2}
\ifx\xfnm\relax \def\xfnm[#1]{\unskip,\space#1}\fi
\bibitem[{Hupkes et~al.(2020)Hupkes, Dankers, Mul, and
  Bruni}]{jair_HupkesDMB20}
\bibinfo{author}{D.~Hupkes}, \bibinfo{author}{V.~Dankers},
  \bibinfo{author}{M.~Mul}, \bibinfo{author}{E.~Bruni},
\newblock \bibinfo{title}{Compositionality decomposed: How do neural networks
  generalise?},
\newblock \bibinfo{journal}{J. Artif. Intell. Res.} \bibinfo{volume}{67}
  (\bibinfo{year}{2020}) \bibinfo{pages}{757--795}. \URLprefix
  \url{https://doi.org/10.1613/jair.1.11674}.
  \DOIprefix\doi{10.1613/jair.1.11674}.
\bibitem[{Mahdavi et~al.(2023)Mahdavi, Swersky, Kipf, Hashemi, Thrampoulidis,
  and Liao}]{mahdavi2023towards}
\bibinfo{author}{S.~Mahdavi}, \bibinfo{author}{K.~Swersky},
  \bibinfo{author}{T.~Kipf}, \bibinfo{author}{M.~Hashemi},
  \bibinfo{author}{C.~Thrampoulidis}, \bibinfo{author}{R.~Liao},
\newblock \bibinfo{title}{Towards better out-of-distribution generalization of
  neural algorithmic reasoning tasks},
\newblock \bibinfo{journal}{Transactions on Machine Learning Research}
  (\bibinfo{year}{2023}). \URLprefix
  \url{https://openreview.net/forum?id=xkrtvHlp3P}.
\bibitem[{Hitzler(2022)}]{hitzler2022neuro}
\bibinfo{author}{P.~Hitzler},
\newblock \bibinfo{title}{Neuro-symbolic artificial intelligence: The state of
  the art}  (\bibinfo{year}{2022}).
\bibitem[{Testolin(2023)}]{testolin2023can}
\bibinfo{author}{A.~Testolin},
\newblock \bibinfo{title}{Can neural networks do arithmetic? a survey on the
  elementary numerical skills of state-of-the-art deep learning models},
\newblock \bibinfo{journal}{arXiv preprint arXiv:2303.07735}
  (\bibinfo{year}{2023}).
\bibitem[{Vaswani et~al.(2017)Vaswani, Shazeer, Parmar, Uszkoreit, Jones,
  Gomez, Kaiser, and Polosukhin}]{Vaswani2017AttentionIA}
\bibinfo{author}{A.~Vaswani}, \bibinfo{author}{N.~Shazeer},
  \bibinfo{author}{N.~Parmar}, \bibinfo{author}{J.~Uszkoreit},
  \bibinfo{author}{L.~Jones}, \bibinfo{author}{A.~N. Gomez},
  \bibinfo{author}{L.~Kaiser}, \bibinfo{author}{I.~Polosukhin},
\newblock \bibinfo{title}{Attention is all you need},
\newblock in: \bibinfo{editor}{I.~Guyon}, \bibinfo{editor}{U.~von Luxburg},
  \bibinfo{editor}{S.~Bengio}, \bibinfo{editor}{H.~M. Wallach},
  \bibinfo{editor}{R.~Fergus}, \bibinfo{editor}{S.~V.~N. Vishwanathan},
  \bibinfo{editor}{R.~Garnett} (Eds.), \bibinfo{booktitle}{Advances in Neural
  Information Processing Systems 30: Annual Conference on Neural Information
  Processing Systems 2017, December 4-9, 2017, Long Beach, CA, {USA}},
  \bibinfo{year}{2017}, pp. \bibinfo{pages}{5998--6008}. \URLprefix
  \url{https://proceedings.neurips.cc/paper/2017/hash/3f5ee243547dee91fbd053c1c4a845aa-Abstract.html}.
\bibitem[{Li and McClelland(2022)}]{li2022systematic}
\bibinfo{author}{Y.~Li}, \bibinfo{author}{J.~McClelland},
\newblock \bibinfo{title}{Systematic generalization and emergent structures in
  transformers trained on structured tasks},
\newblock in: \bibinfo{booktitle}{NeurIPS '22 Workshop on All Things Attention:
  Bridging Different Perspectives on Attention}, \bibinfo{year}{2022}.
  \URLprefix \url{https://openreview.net/forum?id=BTNaKmYdQmE}.
\bibitem[{Brown et~al.(2020)Brown, Mann, Ryder, Subbiah, Kaplan, Dhariwal,
  Neelakantan, Shyam, Sastry, Askell, Agarwal, Herbert-Voss, Krueger, Henighan,
  Child, Ramesh, Ziegler, Wu, Winter, Hesse, Chen, Sigler, Litwin, Gray, Chess,
  Clark, Berner, McCandlish, Radford, Sutskever, and
  Amodei}]{NEURIPS2020_1457c0d6}
\bibinfo{author}{T.~Brown}, \bibinfo{author}{B.~Mann},
  \bibinfo{author}{N.~Ryder}, \bibinfo{author}{M.~Subbiah},
  \bibinfo{author}{J.~D. Kaplan}, \bibinfo{author}{P.~Dhariwal},
  \bibinfo{author}{A.~Neelakantan}, \bibinfo{author}{P.~Shyam},
  \bibinfo{author}{G.~Sastry}, \bibinfo{author}{A.~Askell},
  \bibinfo{author}{S.~Agarwal}, \bibinfo{author}{A.~Herbert-Voss},
  \bibinfo{author}{G.~Krueger}, \bibinfo{author}{T.~Henighan},
  \bibinfo{author}{R.~Child}, \bibinfo{author}{A.~Ramesh},
  \bibinfo{author}{D.~Ziegler}, \bibinfo{author}{J.~Wu},
  \bibinfo{author}{C.~Winter}, \bibinfo{author}{C.~Hesse},
  \bibinfo{author}{M.~Chen}, \bibinfo{author}{E.~Sigler},
  \bibinfo{author}{M.~Litwin}, \bibinfo{author}{S.~Gray},
  \bibinfo{author}{B.~Chess}, \bibinfo{author}{J.~Clark},
  \bibinfo{author}{C.~Berner}, \bibinfo{author}{S.~McCandlish},
  \bibinfo{author}{A.~Radford}, \bibinfo{author}{I.~Sutskever},
  \bibinfo{author}{D.~Amodei},
\newblock \bibinfo{title}{Language models are few-shot learners},
\newblock in: \bibinfo{editor}{H.~Larochelle}, \bibinfo{editor}{M.~Ranzato},
  \bibinfo{editor}{R.~Hadsell}, \bibinfo{editor}{M.~Balcan},
  \bibinfo{editor}{H.~Lin} (Eds.), \bibinfo{booktitle}{Advances in Neural
  Information Processing Systems}, volume~\bibinfo{volume}{33},
  \bibinfo{publisher}{Curran Associates, Inc.}, \bibinfo{year}{2020}, pp.
  \bibinfo{pages}{1877--1901}. \URLprefix
  \url{https://proceedings.neurips.cc/paper/2020/file/1457c0d6bfcb4967418bfb8ac142f64a-Paper.pdf}.
\bibitem[{Ouyang et~al.(2022)Ouyang, Wu, Jiang, Almeida, Wainwright, Mishkin,
  Zhang, Agarwal, Slama, Gray, Schulman, Hilton, Kelton, Miller, Simens,
  Askell, Welinder, Christiano, Leike, and Lowe}]{ouyang2022training}
\bibinfo{author}{L.~Ouyang}, \bibinfo{author}{J.~Wu},
  \bibinfo{author}{X.~Jiang}, \bibinfo{author}{D.~Almeida},
  \bibinfo{author}{C.~Wainwright}, \bibinfo{author}{P.~Mishkin},
  \bibinfo{author}{C.~Zhang}, \bibinfo{author}{S.~Agarwal},
  \bibinfo{author}{K.~Slama}, \bibinfo{author}{A.~Gray},
  \bibinfo{author}{J.~Schulman}, \bibinfo{author}{J.~Hilton},
  \bibinfo{author}{F.~Kelton}, \bibinfo{author}{L.~Miller},
  \bibinfo{author}{M.~Simens}, \bibinfo{author}{A.~Askell},
  \bibinfo{author}{P.~Welinder}, \bibinfo{author}{P.~Christiano},
  \bibinfo{author}{J.~Leike}, \bibinfo{author}{R.~Lowe},
\newblock \bibinfo{title}{Training language models to follow instructions with
  human feedback},
\newblock in: \bibinfo{editor}{A.~H. Oh}, \bibinfo{editor}{A.~Agarwal},
  \bibinfo{editor}{D.~Belgrave}, \bibinfo{editor}{K.~Cho} (Eds.),
  \bibinfo{booktitle}{Advances in Neural Information Processing Systems},
  \bibinfo{year}{2022}. \URLprefix
  \url{https://openreview.net/forum?id=TG8KACxEON}.
\bibitem[{Deshpande et~al.(2021)Deshpande, Chen, and Lee}]{Deshpande2021RecTAR}
\bibinfo{author}{R.~Deshpande}, \bibinfo{author}{J.~Chen},
  \bibinfo{author}{I.~G. Lee},
\newblock \bibinfo{title}{Rect: A recursive transformer architecture for
  generalizable mathematical reasoning},
\newblock in: \bibinfo{booktitle}{International Workshop on Neural-Symbolic
  Learning and Reasoning}, \bibinfo{year}{2021}.
\bibitem[{Setzler et~al.(2022)Setzler, Howland, and
  Phillips}]{DBLP:journals/corr/abs-2201-11766}
\bibinfo{author}{M.~Setzler}, \bibinfo{author}{S.~Howland},
  \bibinfo{author}{L.~A. Phillips},
\newblock \bibinfo{title}{Recursive decoding: {A} situated cognition approach
  to compositional generation in grounded language understanding},
\newblock \bibinfo{journal}{CoRR} \bibinfo{volume}{abs/2201.11766}
  (\bibinfo{year}{2022}). \URLprefix \url{https://arxiv.org/abs/2201.11766}.
  \href{http://arxiv.org/abs/2201.11766}{{\tt arXiv:2201.11766}}.
\bibitem[{Shaw et~al.(2018)Shaw, Uszkoreit, and Vaswani}]{shaw-etal-2018-self}
\bibinfo{author}{P.~Shaw}, \bibinfo{author}{J.~Uszkoreit},
  \bibinfo{author}{A.~Vaswani},
\newblock \bibinfo{title}{Self-attention with relative position
  representations},
\newblock in: \bibinfo{booktitle}{Proceedings of the 2018 Conference of the
  North {A}merican Chapter of the Association for Computational Linguistics:
  Human Language Technologies, Volume 2 (Short Papers)},
  \bibinfo{publisher}{Association for Computational Linguistics},
  \bibinfo{address}{New Orleans, Louisiana}, \bibinfo{year}{2018}, pp.
  \bibinfo{pages}{464--468}. \URLprefix
  \url{https://aclanthology.org/N18-2074}.
  \DOIprefix\doi{10.18653/v1/N18-2074}.
\bibitem[{Trask et~al.(2018)Trask, Hill, Reed, Rae, Dyer, and
  Blunsom}]{NEURIPS2018_0e64a7b0}
\bibinfo{author}{A.~Trask}, \bibinfo{author}{F.~Hill}, \bibinfo{author}{S.~E.
  Reed}, \bibinfo{author}{J.~Rae}, \bibinfo{author}{C.~Dyer},
  \bibinfo{author}{P.~Blunsom},
\newblock \bibinfo{title}{Neural arithmetic logic units},
\newblock in: \bibinfo{editor}{S.~Bengio}, \bibinfo{editor}{H.~Wallach},
  \bibinfo{editor}{H.~Larochelle}, \bibinfo{editor}{K.~Grauman},
  \bibinfo{editor}{N.~Cesa-Bianchi}, \bibinfo{editor}{R.~Garnett} (Eds.),
  \bibinfo{booktitle}{Advances in Neural Information Processing Systems},
  volume~\bibinfo{volume}{31}, \bibinfo{publisher}{Curran Associates, Inc.},
  \bibinfo{year}{2018}. \URLprefix
  \url{https://proceedings.neurips.cc/paper_files/paper/2018/file/0e64a7b00c83e3d22ce6b3acf2c582b6-Paper.pdf}.
\bibitem[{Ruiz et~al.(2021)Ruiz, Ainslie, and
  Onta{\~{n}}{\'{o}}n}]{DBLP:journals/corr/abs-2110-04169}
\bibinfo{author}{L.~Ruiz}, \bibinfo{author}{J.~Ainslie},
  \bibinfo{author}{S.~Onta{\~{n}}{\'{o}}n},
\newblock \bibinfo{title}{Iterative decoding for compositional generalization
  in transformers},
\newblock \bibinfo{journal}{CoRR} \bibinfo{volume}{abs/2110.04169}
  (\bibinfo{year}{2021}). \URLprefix \url{https://arxiv.org/abs/2110.04169}.
  \href{http://arxiv.org/abs/2110.04169}{{\tt arXiv:2110.04169}}.
\bibitem[{Csord{\'a}s et~al.(2021)Csord{\'a}s, Irie, and
  Schmidhuber}]{csordas-etal-2021-devil}
\bibinfo{author}{R.~Csord{\'a}s}, \bibinfo{author}{K.~Irie},
  \bibinfo{author}{J.~Schmidhuber},
\newblock \bibinfo{title}{The devil is in the detail: Simple tricks improve
  systematic generalization of transformers},
\newblock in: \bibinfo{booktitle}{Proceedings of the 2021 Conference on
  Empirical Methods in Natural Language Processing},
  \bibinfo{publisher}{Association for Computational Linguistics},
  \bibinfo{address}{Online and Punta Cana, Dominican Republic},
  \bibinfo{year}{2021}, pp. \bibinfo{pages}{619--634}. \URLprefix
  \url{https://aclanthology.org/2021.emnlp-main.49}.
  \DOIprefix\doi{10.18653/v1/2021.emnlp-main.49}.
\bibitem[{Dehghani et~al.(2019)Dehghani, Gouws, Vinyals, Uszkoreit, and
  Kaiser}]{dehghani2018universal}
\bibinfo{author}{M.~Dehghani}, \bibinfo{author}{S.~Gouws},
  \bibinfo{author}{O.~Vinyals}, \bibinfo{author}{J.~Uszkoreit},
  \bibinfo{author}{L.~Kaiser},
\newblock \bibinfo{title}{Universal transformers},
\newblock in: \bibinfo{booktitle}{International Conference on Learning
  Representations}, \bibinfo{year}{2019}. \URLprefix
  \url{https://openreview.net/forum?id=HyzdRiR9Y7}.
\bibitem[{Csord{\'a}s et~al.(2022)Csord{\'a}s, Irie, and
  Schmidhuber}]{csordas2022ndr}
\bibinfo{author}{R.~Csord{\'a}s}, \bibinfo{author}{K.~Irie},
  \bibinfo{author}{J.~Schmidhuber},
\newblock \bibinfo{title}{The neural data router: Adaptive control flow in
  transformers improves systematic generalization},
\newblock in: \bibinfo{booktitle}{International Conference on Learning
  Representations}, \bibinfo{year}{2022}. \URLprefix
  \url{https://openreview.net/forum?id=KBQP4A_J1K}.
\bibitem[{Dai and Muggleton(2021)}]{ijcai2021p254}
\bibinfo{author}{W.-Z. Dai}, \bibinfo{author}{S.~Muggleton},
\newblock \bibinfo{title}{Abductive knowledge induction from raw data},
\newblock in: \bibinfo{editor}{Z.-H. Zhou} (Ed.),
  \bibinfo{booktitle}{Proceedings of the Thirtieth International Joint
  Conference on Artificial Intelligence, {IJCAI-21}},
  \bibinfo{publisher}{International Joint Conferences on Artificial
  Intelligence Organization}, \bibinfo{year}{2021}, pp.
  \bibinfo{pages}{1845--1851}. \URLprefix
  \url{https://doi.org/10.24963/ijcai.2021/254}.
  \DOIprefix\doi{10.24963/ijcai.2021/254}, \bibinfo{note}{main Track}.
\bibitem[{Manhaeve et~al.(2018)Manhaeve, Dumancic, Kimmig, Demeester, and
  De~Raedt}]{NEURIPS2018_dc5d637e}
\bibinfo{author}{R.~Manhaeve}, \bibinfo{author}{S.~Dumancic},
  \bibinfo{author}{A.~Kimmig}, \bibinfo{author}{T.~Demeester},
  \bibinfo{author}{L.~De~Raedt},
\newblock \bibinfo{title}{Deepproblog: Neural probabilistic logic programming},
\newblock in: \bibinfo{editor}{S.~Bengio}, \bibinfo{editor}{H.~Wallach},
  \bibinfo{editor}{H.~Larochelle}, \bibinfo{editor}{K.~Grauman},
  \bibinfo{editor}{N.~Cesa-Bianchi}, \bibinfo{editor}{R.~Garnett} (Eds.),
  \bibinfo{booktitle}{Advances in Neural Information Processing Systems},
  volume~\bibinfo{volume}{31}, \bibinfo{publisher}{Curran Associates, Inc.},
  \bibinfo{year}{2018}. \URLprefix
  \url{https://proceedings.neurips.cc/paper_files/paper/2018/file/dc5d637ed5e62c36ecb73b654b05ba2a-Paper.pdf}.
\bibitem[{Marcus(2001)}]{Marcus2001}
\bibinfo{author}{G.~F. Marcus}, \bibinfo{title}{The Algebraic Mind},
  \bibinfo{publisher}{The {MIT} Press}, \bibinfo{year}{2001}. \URLprefix
  \url{https://doi.org/10.7551/mitpress/1187.001.0001}.
  \DOIprefix\doi{10.7551/mitpress/1187.001.0001}.

\end{thebibliography}
